# How Many Examples, and Which Ones? A Study of In-Context Learning for Natural Language Inference

**Georgos Dikaios[1], Isidoros Perikos [1,2,3], and Ioannis Hatzilygeroudis[1,2]**

[1] Hellenic Open University, Patras, GR 26335 Greece

[2] Computer Engineering & Informatics Department, University of Patras, Patras, GR 26504 Greece

[3] Computer Technology Institute and Press Diophantus, Patras, GR 26504 Greece

ORCID: G. Dikaios: https://orcid.org/0009-0004-6766-2129,

## Abstract

Natural language inference (NLI) is a fundamental task of natural language processing and crucial for evaluating the strengths and weaknesses of AI methods. Research has mostly focused on encoder or encoder-decoder architectures. However, the field of decoder-only Large Language Models (LLMs) remains underexplored. While prior work has mainly investigated complex sampling and prompting techniques, the effectiveness of a minimalistic prompting approach is yet to be examined in decoder-only LLMs. In this work, we examine and provide a comprehensive benchmark of LLMs (Llama, Mistral, Gemma, and Phi 4) in this domain, across the MNLI, QNLI and SciTail datasets. The performance of these models was evaluated using zero-shot and few-shot prompting strategies with simple and direct prompts. The results of the article indicate that LLMs generally perform exceptionally well when they are provided with a simple, straightforward prompt, challenging the notion that complex prompting techniques are always required for NLI tasks. Additionally, we find that the choice of demonstration label in one-shot prompting significantly affects performance for most models, with entailment-primed examples performing best. Our findings also show that the use of a few representative examples, in the case of few-shot, produced performance comparable or superior to prompts with higher example counts, although the efficacy of the prompting techniques is model and dataset-dependent. Moreover, our simple prompting approach consistently matched or outperformed more elaborate methods such as ICS and MCQ-based prompting.

## 1. Introduction

Natural Language Inference is a cornerstone of Natural Language Processing (NLP) and more specifically for language understanding. Many core NLP tasks, such as question answering, summarization, fact verification, and dialogue systems, rely implicitly on the ability to perform inference, making NLI a fundamental building block for broader language understanding. Furthermore, NLI has been widely adopted as a standard benchmark for assessing reasoning capabilities, as it requires models to go beyond surface-level patterns and interpret meaning at a deeper level [1]. Its importance extends beyond evaluation, as NLI can also serve as a general framework for reformulating various classification tasks, and also enable flexible zero-shot and few-shot learning across domains [2].

In general, the task of NLI is to determine whether or not one sentence, called the hypothesis, is logically derived from another sentence, called the premise [3]. This relationship is typically categorized into three classes: entailment, where the hypothesis follows from the premise; contradiction, where the hypothesis conflicts with the premise; and neutral, where no clear logical relationship can be established. The task requires models to capture not only syntactic structure but also deeper semantic meaning, contextual nuances, and implicit knowledge [4]. As a result, NLI is considered a challenging benchmark that tests a model's ability to perform reasoning rather than simple pattern matching, making it a key indicator of true language understanding [20]. Motivated by the importance of such semantic relationships in human everyday language, models have been tasked to identify this relationship between the premise and hypothesis.

Many studies before have used encoder-only models for this purpose, namely BERT [5], RoBERTa [6], and DeBERTa [7], showcasing the exceptional performance of such model architectures and even outperforming humans in certain tasks in the case of DeBERTa. As far as decoder-only models are concerned, the applicability remains highly important. Instead of being trained with NLI as the main task, modern LLMs are now using it as a way to judge the models' performance and hallucination occurrences.

Contrary to earlier transformer-based models being evaluated on NLI tasks such as MNLI [8] and QNLI [9], contemporary LLMs are now focusing on more general reasoning benchmarks, such as MMLU [10]. Consequently, reports of specific NLI datasets on performance are scarce, making the evaluation of these models even more necessary. Additionally, there is a gap of reports on performance metrics for more specialized datasets which require scientific understanding, like SciTail [11]. Prior work that has used modern LLMs on MNLI has successfully increased the performance of LLMs using different sampling strategies [12] and prompt-engineering techniques [13, 14], while inspecting the effects of zero-shot and few-shot prompting.

As LLMs become increasingly capable, the temptation to design elaborate prompting pipelines has grown, often at the cost of simplicity and interpretability. Despite these advancements, there is still no clear consensus on the most effective prompting strategy for NLI tasks, especially for small and efficient LLMs [15], with a more empirical approach usually being the default. Existing approaches often prioritize increasingly complex prompt designs or larger numbers of in-context examples, assuming that more information leads to better performance [16]. However, this assumption does not always hold, as overly complex prompts may introduce noise, increase computational overhead, and potentially distract the model from the core task [17]. Additionally, the effectiveness of these techniques appears to be highly dependent on the model architecture, dataset characteristics, and task formulation, leaving an open question as to whether simpler and more direct prompting strategies could achieve comparable or even superior results. This question is particularly relevant for small-scale open-weight models, which are increasingly deployed in resource-constrained settings and have received comparatively less systematic attention in NLI evaluations.

In this work, we argue that simple and direct prompts, without added complexity, are sufficient for small-scale models to achieve competitive performance on NLI tasks. We believe that the simplicity of our prompts allows the models to focus on the task at hand in both zero-shot prompting and few-shot prompting, without confusing it. To validate this hypothesis, we evaluated five open-weight instruction-tuned models: Mistral-7B-Instruct-v0.3 [18], Gemma-2-9B-Instruct [19], Llama-3.1-8B-Instruct, Llama-3.2-3B-Instruct [20], and Phi-4-Mini [21]. These models were selected to represent a diverse yet comparable set of small-scale, instruction-tuned architectures, enabling a systematic comparison across different training strategies and model families. These models were evaluated using the MNLI, QNLI, and SciTail datasets, that cover general domains, question answering using Wikipedia as the source of the questions, as well as scientific questions taken from science exams.

In this context, our study addresses four main research questions: RQ1: How do modern instruction-tuned decoder-only LLMs perform on traditional NLI tasks and how do they compare to encoder-only and encoder-decoder models in zero-shot and few-shot settings? RQ2: How exactly does the label of the demonstration example affect performance in one-shot prompting? RQ3: How exactly does a higher number of in-context demonstration examples affect performance across models? RQ4: How does our prompting approach compare to other prompting techniques from literature?

The main contributions of this work are as follows. First, we provide a comprehensive benchmark of modern decoder-only LLMs on traditional NLI tasks, addressing the lack of systematic evaluation on datasets such as MNLI, QNLI, and SciTail. Unlike prior studies that primarily focus on encoder-based architectures or large-scale prompting pipelines, our work evaluates small and medium-sized open-weight LLMs under a unified and controlled prompting framework. Second, we analyse the effect the demonstration examples have in one-shot, showing different results for different model – dataset combinations. Third, we systematically analyse the impact of zero-shot and few-shot prompting, showing that a small number of demonstration examples is often sufficient and, in many cases, more effective than larger example sets. The results show that a few, indicative of each class, examples perform as well as higher numbers of demonstration examples. This finding indicates that prompt efficiency plays a critical role in practical LLM deployment. Fourth, we provide a comparative analysis of how different model architectures respond to prompt simplicity across datasets of varying complexity and domain characteristics. We demonstrate that simple and direct prompting strategies can achieve performance comparable to, or even surpassing, more complex prompt-engineering techniques, challenging the prevailing assumption that complexity is necessary for strong performance. Finally, we highlight the capability of small-scale open-weight models to perform competitively on NLI tasks, providing insights into efficient and reproducible approaches for evaluating and deploying LLMs. To the best of our knowledge, no prior work systematically benchmarks small-to-medium instruction-tuned decoder-only LLMs on natural language inference tasks under a prompting framework with logit-based label scoring.

In the remainder of the paper we review the relevant literature and related work in Section 2 and present the selected model architectures used in this study in Section 3. In Section 4 we present the methodology, including the datasets, models, and evaluation setup. In Section 5 we present and analyse the experimental results, highlighting the performance of each model under both zero-shot and few-shot prompting settings, along with the intricacies of the demonstration example selection. Finally, Sections 6 and 7 conclude the paper by recognising limitations, summarizing the key results and outlining directions for future work.

## 2. Related Work

In recent years, NLI has attracted significant attention due to its central role in evaluating language understanding and reasoning [2]. A large body of research focuses on the development of methods capable of accurately identifying semantic relationships between sentences and efficiently perform inference and reasoning tasks [22, 23]. Several studies have explored how prompts affect the performance of large language models in various question answering and inference procedures [24–26]. While prior work has explored prompt engineering strategies for improving performance in natural language inference, most studies either focus on large-scale models or employ complex prompting techniques such as Chain-of-Thought (CoT) reasoning, in-context sampling, or prompt ensembling. As a result, there is limited understanding of how simpler prompting strategies affect the performance of small and medium-sized decoder-only models across multiple NLI benchmarks in a consistent evaluation setting. This section reviews prior work on NLI and prompting strategies, with a particular focus on how different approaches influence model performance.

### *2.1 Encoder-only Models*

A lot of work has been done using encoder-only models, achieving very high accuracy, making them a gold standard for NLI tasks. These models rely on bidirectional self-attention, allowing them to capture complex semantic relationships between words and phrases, which is essential for accurately modelling entailment and contradiction. In 2018 the first well-recognized encoder-only model, BERT [5], was published showcasing the effectiveness of encoder-only models. Two models were released, BERTBASE and BERTLARGE, achieving an accuracy after fine-tuning on MNLI of 84.6% and 86.7% respectively and on QNLI of 90.5% and 92.7% respectively [5].

A year later RoBERTa [6] and ERNIE 2.0 Base and Large [27] were released, both beating BERT. RoBERTa was a direct improvement upon BERT, increasing the scale of the training and reaching an accuracy on MNLI and QNLI of 90.2% and 94.7% respectively [6]. At the same time, ERNIE 2.0 also outperformed BERT using a continual multi-task framework. This allowed the model to acquire a better grasp of lexical, syntactic, and semantic logic. ERNIE 2.0 Base reached an accuracy on MNLI and QNLI of 86.1% and 92.9% respectively, while ERNIE 2.0 Large had 88.7% and 94.6% [27]. These improvements highlighted the importance of large-scale pretraining and richer training objectives, which contributed to better generalization and a deeper understanding of linguistic structures.

Finally, the last major breakthrough in NLI came with the release of DeBERTa. This model's architecture fundamentally changed how attention works using disentangled attention for the logic of word relationships. DeBERTa Base reached an accuracy of 88.8% on MNLI and DeBERTa Large 91.1%. Overall, encoder-only models have set a strong baseline for NLI performance, often approaching or surpassing human-level accuracy on benchmark datasets. However, their reliance on supervised fine-tuning and task-specific architectures limits their flexibility compared to more recent general-purpose language models.

## *2.2 Encoder-Decoder Models*

While previous models used a classification head, in 2019 the T5 model was released, introducing the "text-to-text" framework [28]. This approach reformulated all NLP tasks, including NLI, into a unified text generation problem, allowing the model to generate the target label as text rather than predicting it through a classification layer. This design enabled greater flexibility and transferability across tasks, as the same architecture and training objective could be applied universally. This new approach resulted in an accuracy of 92.2% on MNLI and of 96.9% on QNLI [28]. Later, the Flan-T5 model [29] was introduced, which was fine-tuned to follow instructions on a substantially larger collection of tasks, including CoT examples. Studies using this model for NLI tasks report an accuracy as high as 70.1% on the MNLI dataset [12]. Encoder-decoder models remain largely confined to the T5 family, due to recent work favouring decoder-only LLMs for their efficiency, scalability, and strong performance.

## *2.3 Decoder-only Models*

Research on traditional NLI tasks is sparse for decoder-only models. The focus has mostly shifted to general knowledge and understanding tasks like MMLU, neglecting the performance reports on datasets such as MNLI, which have been the cornerstone for evaluating BERT-like models.

The scalability, flexibility, and ability of decoder only models to operate in zero-shot and few-shot settings make them highly attractive for practical applications. This creates a strong motivation for systematically evaluating their capabilities on NLI tasks and understanding how factors such as prompt simplicity, number of examples, and dataset characteristics influence their performance.

A recent study examined the relevance of traditional NLI tasks for encoder-only models. They concluded that they are still very useful and able to discriminate between models of different qualities [30]. In their study they used a multiple-choice prompting style and evaluated the Llama and Mistral base model families of various sizes. The model was given sample examples chosen randomly and then the question with the labels as options and was asked to predict the correct letter choice. The results that they report on the MNLI dataset that can be most closely related to our study are for Llama 3.1 8B (~67% Accuracy) and Mistral 7B (~67% Accuracy).

Despite the growing adoption of decoder-only models, their evaluation on traditional NLI benchmarks remains limited and fragmented. Existing studies often report results on a single dataset or focus on specific prompting strategies, making it difficult to draw general conclusions about their behaviour. A comprehensive and consistent benchmark across multiple datasets and prompting configurations is still lacking.

### *2.4 Approaches to In-Context Learning*

There have been many studies which have tried various techniques for improving in-context learning (ICL). ICL is defined as the use of the text input as a form of task specification [31]. This includes the instructions in zero-shot and few-shot setups as well as the demonstration examples provided in few-shot.

A recent study focused on the selection process of the examples being shown to the model when few-shot prompting, introducing in-context sampling as a method [12]. Using this strategy, they were able to reach an accuracy of 63.4% for k=20 examples on Mistral 7B.

Although several studies [31], [32] suggest that increasing the number of in-context examples improves performance, more recent findings [33] indicate that this trend does not consistently hold, especially for smaller models. In particular, excessive examples may lengthen the input and introduce noise, dilute the informative demonstrations, and negatively affect model performance [33, 34]. However, these observations have not been systematically validated across multiple NLI datasets and modern instruction-tuned LLMs.

In summary, while prior work has explored various prompting techniques and model architectures for NLI, there is still no clear consensus on the exact effectiveness of prompting strategies, particularly for smaller instruction-tuned LLMs. Moreover, there is a gap, to the best of our knowledge, when it comes to the selection of demonstration examples in one-shot context for NLI. This work addresses these gaps by providing a systematic evaluation of prompt simplicity across multiple datasets and models, without relying on advanced prompt engineering techniques, while examining how the choice of demonstration example affects one-shot ICL.

## 3. Model Architectures

This section discusses the models selected for our study and their architecture. We are using five contemporary, instruction-tuned LLMs that range from small to medium sizes. Unlike earlier base versions, these models have undergone supervised fine-tuning and reinforcement learning from human feedback (RLHF) to increase their performance when prompted to follow direct instructions, especially with zero-shot prompts.

The Llama family utilizes a standard dense transformer architecture [20]. Llama 3.1 (8B) Instruct was pre-trained on a corpus of over 15 trillion tokens and features a 128k tokens context window and was later refined through knowledge distillation (KD) from the 405B parameter version. It also features grouped-query attention to maintain a high inference speed. Llama 3.2 (3B) Instruct was developed using structured single-shot pruning from Llama 3.1 8B, followed by KD from both the larger Llama 3.1 8B and 70B models. This allows the model to keep the complex reasoning found in the larger Llama 3.1 models making it ideal for testing simple zero-shot and few-shot effectiveness [35].

Mistral 7B [18] is widely recognized for its efficiency. Mistral is based on a transformer architecture, utilizes the sliding window attention (SWA) to manage memory, allowing it to handle big sequences with a reduced inference cost by restricting the attention span to a fixed window, and like the Llama family, it also utilizes the grouped-query attention. The Instruct v0.3 version has an extended vocabulary size and context window of 32,768 tokens, and supports the v3 tokenizer, which optimizes the model for instruction following and complex tasks.

Gemma 2 9B [19] is based on a transformer architecture and introduces alternating local and global attention layers, where local attention also uses the SWA mechanism and global attention operates over the full context. It also features grouped-query attention for faster inference and reduced memory usage. Additionally, the 9B model is trained using KD from the 27B model instead of standard next-token prediction. This model is the largest one we will use as an upper limit, making it ideal for comparison to the smaller ones.

Phi-4-Mini is a 3.8B-parameter dense decoder-only transformer. It also uses grouped-query attention and shares the input and output embeddings, and compared to its predecessor Phi-3.5-Mini it expands the vocabulary to 200k tokens to better support multilingual use. The model supports a context window of 128k tokens and is trained on 5T tokens, with an emphasis on reasoning-dense content such as math and coding. Despite its compact size it is reported to match models twice its size on tasks that require complex reasoning, which makes it an interesting point of comparison for the other models we use.

## 4. Methodology

This section presents the overall work methodology and describes the task formulation, the datasets used, and the main aspects of the evaluation setup. The experimental design aims to properly examine the effect of prompt simplicity and example quantity on model performance.

### *4.1 Task Formulation and Datasets*

The datasets used in our study are the MNLI [8], QNLI [9] and SciTail [11], covering different levels of complexity and allowing a more comprehensive evaluation of model behaviour under varying inference conditions. MNLI and QNLI have been a cornerstone for benchmarking NLI, and with the addition of SciTail we cover a wide range of domains including specialized scientific questions. The three datasets differed in difficulty; QNLI is widely considered an easier dataset, because of its binary and question answering nature, while SciTail and MNLI are generally considered harder. SciTail, even though it is also a binary problem, it requires the understanding of scientific notions, while MNLI is a ternary problem, that introduces the concept of contradiction explicitly. Whereas in the cases of QNLI and SciTail the model only needs to classify based on the existence or absence of sufficient proof, for MNLI it also has to differentiate between neutrality and contradiction. The selected datasets allow the evaluation of the models' functionality under both binary and multi-class semantic inference frameworks, that range from general linguistic reasoning to domain-specific scientific understanding. For our work, we used the pre-split validation set for the MNLI and QNLI datasets and the pre-split test set for the SciTail dataset.

The problem was formulated as a pair of sentences $(P, H)$, where $P$ is the premise and $H$ is the hypothesis. The labels were mapped from their categorical indices $\{0, 1\}$ for QNLI and from $\{0, 1, 2\}$ in the case of MNLI to text and the label space $Y_d$ is defined as:

$$Y_{MNLI} \in \{Entailment, Contradiction, Neutral\} \quad (1)$$

$$Y_{QNLI} \in \{Entailment, Not\ Entailment\} \quad (2)$$

$$Y_{SciTail} \in \{Entails, Neutral\} \quad (3)$$

The SciTail dataset already had the labels as "entails" and "neutral" and we kept them as such. We decided however, to use "Not Entailment" instead of "Not_Entailment" as the label mapping for QNLI, to follow more closely the natural way of writing.

**Table 1** Representative examples of Premise-Hypothesis

| Dataset | Premise/Question | Hypothesis/Sentence | Label |
|---|---|---|---|
| *MNLI* | Clearly the press has done a lousy job with its focus on behavior such as infidelity or drug use that most people don't care about. | The press has focused on various topics that most readers aren't interested in. | Entailment |
| | The family. | The couple. | Neutral |
| | I knew him and liked and respected him. | I never knew the man. | Contradiction |
| *QNLI* | Where are some of the best medical facilities in East-Central Europe located? | Today, Warsaw has some of the best medical facilities in Poland and East-Central Europe | Entailment |
| | What land was ceded to Spain? | France ceded its territory east of Mississippi to Great Britain | Not Entailment |
| *SciTail* | This causes gasses to become liquids and liquids to become solids as the temperature decreases. | Gasses and liquids become solids at low temperatures | Entails |
| | The mud layers formed by the slow continuous rain of finer sediment settling from the ocean | When sediments settle out of water, the form horizontal layers | Neutral |

Table 1 showcases representative examples of each label $Y_d$, taken from the subsets we created for the few-shot evaluation and were provided to the models during prompting. Both the MNLI and QNLI datasets were completely balanced, whereas the SciTail dataset was not with 63.3% of the labels being “Neutral”. For this reason, when we were pooling examples for few-shot, we performed stratified sampling to respect this imbalance and get a subset representative to the original dataset and maintain the class distribution. Pooling was performed with random examples taken from the datasets without replacement, which were then removed, so as not to have any data leakage during evaluation, and saved to NumPy arrays to ensure reproducibility.

For few-shot evaluation we tried different numbers of examples. In the cases of QNLI and SciTail, we started with one-shot, 1 example per class, then 2 examples – one of each class in $Y_{QNLI}$ and $Y_{Scitail}$ so that they are both represented – and then the number of demonstration examples provided for a single subset seed was in [3, 5, 8, 10, 15] for each seed. Similarly, for MNLI, to accommodate for the different number of classes $Y_{MNLI}$, we started again with one-shot, 1 example per class, and then 3 examples – one of each class in $Y_{MNLI}$ – and then similarly, the number of demonstration examples given was in [5, 8, 10, 15].

## *4.2 Model Selection and Configuration*

We selected to use models small in size for the evaluation because we find it very important to see the performance of small LLMs that have lower environmental impact, smaller computational costs and ease of access for reproducibility reasons. With that in mind, we used Mistral-7B-Instruct-v0.3, Gemma-2-9B-Instruct, Llama-3.1-8B-Instruct, Llama-3.2-3B-Instruct, and Phi-4-mini-instruct, performing 8-bit quantization with the BitsAndBytes library in order to address the size of these models. Llama-3.2 and Phi-4, being the smallest, were considered a baseline for all other models to beat.

To assess robustness to the choice of demonstration examples, we sampled 10 random subsets of 15 demonstration examples each (10 seeds), for a total of 150 examples, without replacement. These 150 examples were then excluded from the original dataset, thus creating a new, clean evaluation set, which was then used for all evaluations. The same remaining evaluation set was used for all prompting conditions.

During the prompting we used the chat template for each model and the experiments were conducted on an A100 GPU. The input was also truncated to max length, which was calculated for each setup and prompt size. In order to not lose a lot of information and affect performance, while we regulate the size of the prompts, we only trimmed 0.2% of them. The other 99.8% had their lengths set to max length with padding from the left side to ensure that the next token generated will continue from the prompt. Batch size used was 32 for both zero-shot and few-shot prompting.

The prompt used was similar for all datasets with a few differentiations to accommodate for the changes in nature for each dataset. The prompt we used for zero-shot was the following:

```
Does the premise entail the hypothesis? Answer exactly one word in lowercase: {label_options}.
Premise: {premise}
Hypothesis: {hypothesis}
Answer:
```

and the prompt we used for the few-shot was the following:

```
Example: {n}:
Premise: {premise_n}
```

```
Hypothesis: {hypothesis_n}
Answer: {label_n}

[repeated for n=1, …, k demonstrations]

Given the above examples as reference does the premise entail the hypothesis in the following case? Answer exactly one word in lowercase as in the examples: {label_options}.

Premise: {premise}
Hypothesis: {hypothesis}
Answer:
```

For QNLI "premise" was substituted with "question" and "hypothesis" with "sentence". The prompts were designed to be simple, direct and avoid additional instructions or reasoning cues that could introduce unnecessary complexity. Also, no additional reasoning instructions, decomposition steps, or any auxiliary contextual cues were used so as to allow the evaluation to focus solely on the effects of prompt simplicity and quantity. We used a fixed prompt formulation for all models within each dataset, allowing us to compare the models in the same way.

## *4.3 Evaluation Metrics*

Many studies report only accuracy as a metric for performance, in our study however we found it very important to report not only the Accuracy as the model's main evaluation of performance but also macro F1-score, and Matthews Correlation Coefficient (MCC). While Accuracy and F1-Score monitor the overall performance of the model, and MCC provide additional meaningful measure of the correlation between the predicted class and the ground truth and are defined as such:

$$MCC = \frac{c \times s - \sum_k^K p_k \times t_k}{\sqrt{(s^2 - \sum_k^K {p_k}^2) \times (s^2 - \sum_k^K {t_k}^2)}} \qquad (5)$$

Where:

- c: Total number of correctly predicted samples
- s: Total number of samples
- $t_k$: Total number of times class k was predicted
- $p_k$: Total number of times class k occurred in the ground truth

## *4.4 Prediction Extraction*

Prediction extraction was done through logits instead of stochastic text generation by computing the conditional log-probability of each candidate label sequence, given the input prompt.

For each target label $y_d$ in $Y_d$ let $Y_{d,t}$ be the t-th token in the sequence. The conditional log-probability of the label $Y_d$ given the input X is calculated by summing the log-probabilities of each token, given the input and the preceding tokens in the label:

$$L(Y_d|X) = \sum_{t=1}^{|Y_d|} log\, P(Y_{d,t}|X, y_{d,1}, \dots, y_{d,t-1}) \in (-\infty, +\infty) \qquad (6)$$

where:

- $X$: input/prompt
- $Y_d$: Target label sequence
- $Y_{d,t}$: t-th token of the label $Y_d$

To ensure the scores are comparable across labels of different sizes and to convert them into valid probabilities, we normalize the raw log probabilities using the $LogSumExp$. Therefore, the normalized probability P(Yd|X) for label yd in Yd is defined as:

$$P(Y_d|X) = \frac{exp\ (L(Y_d|X))}{\sum_{j\in Y_d} exp\ (L(Y_j|X))} \epsilon\ [0,1] \qquad (7)$$

and the final predicted label yd is then selected by taking the maximum argument of the normalized probabilities:

$$y_d = arg \max_{k\ \epsilon\ Y_d} P(Y_d|X) \quad (8)$$

The candidate labels may contain different numbers of tokens something that depends on the tokenizer of each model.

## 5. Results

This section presents the experimental results and analyzes the impact of prompt design and example quantity on model performance both in zero-shot and few-shot (Figure 1, Table 2). Moreover, we are looking at the effect the demonstration examples have in one-shot and provide our insights as to why (Tables 3-5). We then present each model's performance across different number of in-context demonstration examples and identify which prompt setup produced the best result (Figures 2-4, Table 6). We then conclude with a comparison of performance across different models and configurations.

### *5.1 Overall Performance*

In this subsection we provide an overview of the performance for each model/dataset combination. In Figure 1 we present a comparison of the best F1-scores the models were able to achieve in each dataset.

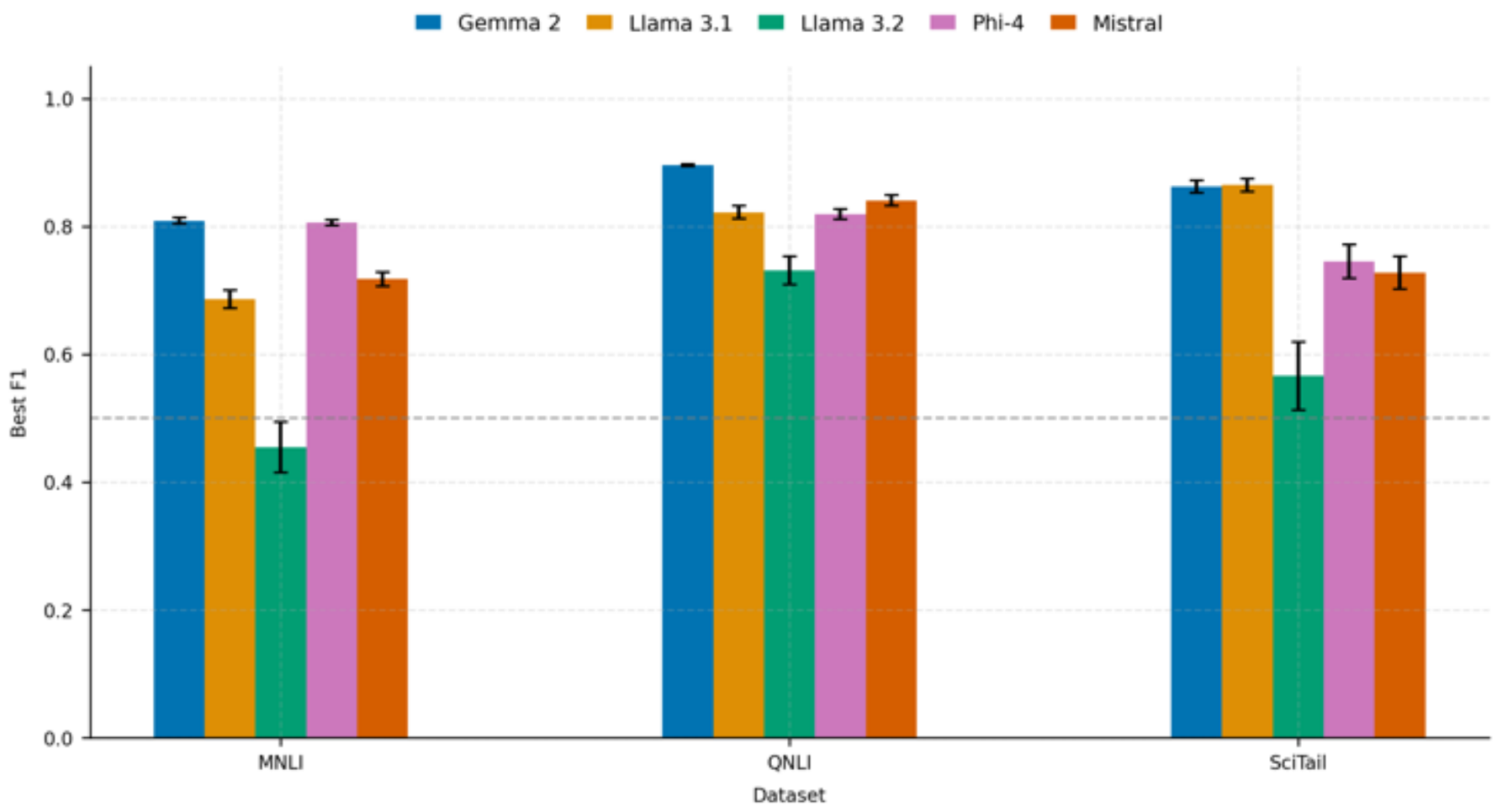

**Fig. 1** A comparison of performance, showing the best F1-Score per model and dataset with 95% CIs

The detailed results of the models reporting Accuracy, F1-Score, and MCC for each model per dataset along with the corresponding 95% CIs can be found in Table 2. The confidence intervals were computed as the mean ± t * SEM, where SEM is the standard error of the mean across seeds and t is the critical value of the t-distribution at the 95% confidence level with n-1 degrees of freedom. We see that the models generally performed better on the QNLI dataset supporting the claim that it is the easier task in our pipeline. Llama 3.2 clearly had the worst performance as indicated by low metrics across the board. The rest of the models had very similar results with Gemma showcasing the greatest consistency and predictive power. This was to be expected and further reinforces the fact that the size of a model's parameters plays one of the most important roles for a model's performance [5, 31], [36]. Two cases stand out as exceptions; Phi-4 outclassed Llama 3.2 in all tasks and in some cases, it even performed better than mistral and Llama 3.1, validating its strong performance claims on reasoning tasks, despite its small size and Llama 3.1 outperformed all other models on SciTail. The Llama family has a very unstable performance, making it hard to evaluate whether this performance of the models comes from their training to specific tasks or from data leakage during training.

**Table 2** Best Results per Model and Dataset. Average across 10 seeds, with 95% CIs

| Model | Dataset | Accuracy | F1-Score | MCC |
|---|---|---|---|---|
| Llama 3.2 | QNLI | 0.737 ± 0.019 | 0.731 ± 0.022 | 0.495 ± 0.027 |
| | MNLI | 0.504 ± 0.031 | 0.454 ± 0.040 | 0.290 ± 0.035 |
| | SciTail | 0.614 ± 0.041 | 0.566 ± 0.054 | 0.209 ± 0.047 |
| *Llama 3.1* | QNLI | 0.8231 ± 0.0097 | 0.822 ± 0.010 | 0.653 ± 0.016 |
| | MNLI | 0.689 ± 0.014 | 0.686 ± 0.014 | 0.547 ± 0.018 |
| | **SciTail** | **0.8712 ± 0.0094** | **0.865 ± 0.010** | **0.735 ± 0.016** |
| *Mistral* | QNLI | 0.8411 ± 0.0079 | 0.8406 ± 0.0083 | 0.687 ± 0.013 |
| | MNLI | 0.722 ± 0.012 | 0.718 ± 0.011 | 0.587 ± 0.017 |
| | SciTail | 0.728 ± 0.025 | 0.728 ± 0.025 | 0.526 ± 0.041 |
| *Gemma 2* | **QNLI** | **0.8959 ± 0.0017** | **0.8959 ± 0.0017** | **0.7921 ± 0.0033** |
| | **MNLI** | **0.8086 ± 0.0048** | **0.8089 ± 0.0046** | **0.7172 ± 0.0062** |
| | SciTail | 0.864 ± 0.010 | 0.8618 ± 0.0097 | 0.737 ± 0.014 |
| *Phi 4* | QNLI | 0.8197 ± 0.0075 | 0.8189 ± 0.0077 | 0.647 ± 0.014 |
| | MNLI | 0.8074 ± 0.0047 | 0.8059 ± 0.0047 | 0.7137 ± 0.0065 |
| | SciTail | 0.746 ± 0.026 | 0.745 ± 0.026 | *0.5575 ± 0.030* |

## *5.2 Demonstration Label Effect*

In Table 3 we can see the models' performance qualitatively during one shot ICL. The results show a clear pattern; the demonstration examples that have the entailment label give the best results in most cases. Mistral is the only model that doesn't show this trend.

**Table 3** Best performance of the models during one-shot for each dataset, based on the label of the demonstration example.

| Model | Dataset | Best One-shot Condition |
|---|---|---|
| Llama 3.2 | QNLI | Entailment |
| | MNLI | Entailment |
| | SciTail | Entails |
| *Llama 3.1* | QNLI | Entailment |
| | MNLI | Entailment |
| | SciTail | Entails |
| *Mistral* | QNLI | Not Entailment |
| | MNLI | neutral |
| | SciTail | Neutral |
| *Gemma 2* | QNLI | Entailment |
| | MNLI | Entailment |
| | SciTail | Entailment |
| *Phi 4* | QNLI | Entailment |
| | MNLI | Contradiction |
| | SciTail | Entails |

These results required a further inspection to see if the changes of the performance when the model was given a demonstration example with "entailment" as the label vs other was statistically significant. To test this, we performed a Wilcoxon test for the QNLI and SciTail and then a Friedman chi square test for MNLI, where we then followed with a Wilcoxon test when the omnibus was found significant.

**Table 4** Comparison of one-shot performance, examining the effect of the label the demonstration example has. When omnibus was not significant, we report that and not the pair-wise tests

| Model | Dataset | Comparison | p-value | Effect Size |
|---|---|---|---|---|
| | QNLI | Entailment vs Not Entailment | .203 | 0.4 |
| | **MNLI** | **Contradiction vs Entailment** | **.015** | **0.89** |
| Llama 3.2 | MNLI | Contradiction vs Neutral | .015 | 0.85 |
| | **MNLI** | **Entailment vs Neutral** | **.015** | **0.89** |
| | SciTail | Entails vs Neutral | .260 | 0.36 |
| | **QNLI** | **Entailment vs Not Entailment** | **.005** | **0.89** |
| | **MNLI** | **Contradiction vs Entailment** | **.021** | **0.85** |
| *Llama 3.1* | MNLI | Contradiction vs Neutral | .033 | 0.76 |
| | **MNLI** | **Entailment vs Neutral** | **.047** | **0.63** |
| | SciTail | Entails vs Neutral | .074 | 0.56 |
| | QNLI | Entailment vs Not Entailment | .646 | 0.15 |
| *Mistral* | MNLI | omnibus | .150 | No |
| | SciTail | Entails vs Neutral | .445 | 0.24 |
| *Gemma 2* | QNLI | Entailment vs Not Entailment | .059 | 0.60 |
| | **MNLI** | **Contradiction vs Entailment** | **.015** | **0.89** |
| | MNLI | Contradiction vs Neutral | .646 | 0.15 |
| | **MNLI** | **Entailment vs Neutral** | **.015** | **0.89** |
| | **SciTail** | **Entails vs Neutral** | **.047** | **0.63** |
| | QNLI | Entailment vs Not Entailment | .203 | 0.40 |
| | MNLI | Contradiction vs Entailment | 0.959 | 0.02 |
| *Phi 4* | MNLI | Contradiction vs Neutral | .015 | 0.89 |
| | **MNLI** | **Entailment vs Neutral** | **.015** | **0.89** |
| | **SciTail** | **Entails vs Neutral** | **.005** | **0.89** |

One case clearly stands out, Mistral. This model's performance doesn't seem to be affected by the demonstration examples, with none of the tests providing a statistically significant result. The results for the rest are quite different and mixed. More specifically, we see a lot of cases where the demonstration example mattered, especially for the MNLI dataset. In our analysis we are considering the cases where entailment is compared to another label, as those were the cases the models performed best as is shown in Table 3. Moreover, we are not considering the case of Phi-4 on the MNLI dataset as it showed extreme performance drop in one-shot for MNLI (Figure 2) and the predictions collapsed to a single class (see Appendix A).

In order to further investigate this behavior, we looked at the total variation distance (TVD) between the predicted and true class distributions, pooling predictions across all seeds for a given combination, and we compare the TVD on zero-shot as a baseline and one-shot. The results in Table 5 along the F1 scores in Figures 2, 3 and 4 show that the demonstration example matters for these models. While entailment-primed demonstration examples consistently had the lowest TVD among the one-shot conditions, this did not always translate to overall better calibration over the zero-shot baseline. For Llama 3.2 and Gemma 2, zero-shot achieved a lower TVD than any one-shot condition, including entailment, on MNLI.

**Table 5** Total variation distance between predicted and true class distribution

| Model | Dataset | Zero-Shot | Entailment/ Entails | Contradiction | Not Entailment/ Neutral |
|---|---|---|---|---|---|
| *Llama 3.2* | MNLI | 0.260 | **0.278** | 0.320 | 0.607 |
| *Llama 3.1* | QNLI | 0.236 | **0.139** | 0.242 | 0.224 |
| | MNLI | 0.141 | **0.007** | - | 0.296 |
| *Gemma 2* | MNLI | 0.014 | **0.062** | 0.074 | 0.122 |
| | SciTail | 0.117 | **0.072** | - | 0.089 |
| *Phi 4* | SciTail | 0.361 | **0.183** | - | 0.313 |

### *5.3 Shot Count Performance*

In Table 6 we demonstrate where the best results shown in Table 2 derive from. More specifically, we report the prompt setup (zero-shot, one-shot or few-shot), in the case of few-shot, the number of examples and when there was only one example, what kind of example the model was given.

**Table 6** Best Prompt Setup per Model and Dataset based on F1-Score

| Model | Dataset | ICL Setting - Number of Examples |
|---|---|---|
| Llama 3.2 | QNLI | Few-Shot - 8 Examples |
| | MNLI | One-Shot (Entailment) |
| | SciTail | Few-Shot - 15 Examples |
| *Llama 3.1* | QNLI | One-Shot (Entailment) |
| | MNLI | One-Shot (Entailment) |
| | SciTail | One-Shot (Entails) |
| *Mistral* | QNLI | Few-Shot - 3 Examples |
| | MNLI | Few-Shot - 10 Examples |
| | SciTail | Few-Shot - 2 Examples |
| *Gemma 2* | QNLI | Few-Shot - 15 Examples |
| | MNLI | Few-Shot - 15 Examples |
| | SciTail | Few-Shot - 10 Examples |
| *Phi 4* | QNLI | Few-Shot - 10 Examples |
| | MNLI | Few-Shot - 15 Examples |
| | SciTail | Few-Shot - 5 Examples |

The results at first glance are quite mixed and there is a model dependency. The prompt setup that gave the best results in most cases was few-shot with 10 or 15 examples. Two cases stand out, Gemma 2 and Llama 3.1. On the one hand, Gemma 2 shows best performance with a high count of demonstration examples, and on the other hand Llama 3.1 performed best with one-shot. Although we have reported the best results in Table 2, it is also important to note that in some cases, the second-best results, which were only marginally worse, came from very few examples.

Our case is further reinforced by Figures 2-4. Two clear trends for Gemma 2 and Llama 3.2 are indicated. The first is very consistent throughout all number of examples, almost reaching the performance of state-of-the-art models like DeBERTa Large, while the latter underperforms due to the lower number of parameters and is only doing slightly better than random guessing. Another comparison is between Mistral and Llama 3.1. Although Llama 3.1 has marginally more parameters (1B difference), Mistral's use of SWA compensates for this difference, allowing it for more efficient sequence processing and thus resulting in a stronger overall performance. However, Llama 3.1 significantly overperforms on the SciTail dataset with only a single demonstration example, implying that it has a stronger scientific understanding.

Consistent with the analysis of the demonstration example in section B, models primed with an entailment example generally showed best performance among the one-shot conditions, although this effect was model dependent including the exception of Mistral, which showed no significant sensitivity to the demonstration condition. In general, Llama 3.2 shows the widest confidence intervals across all datasets and the lowest performance, indicating a limitation in the model's capacity.

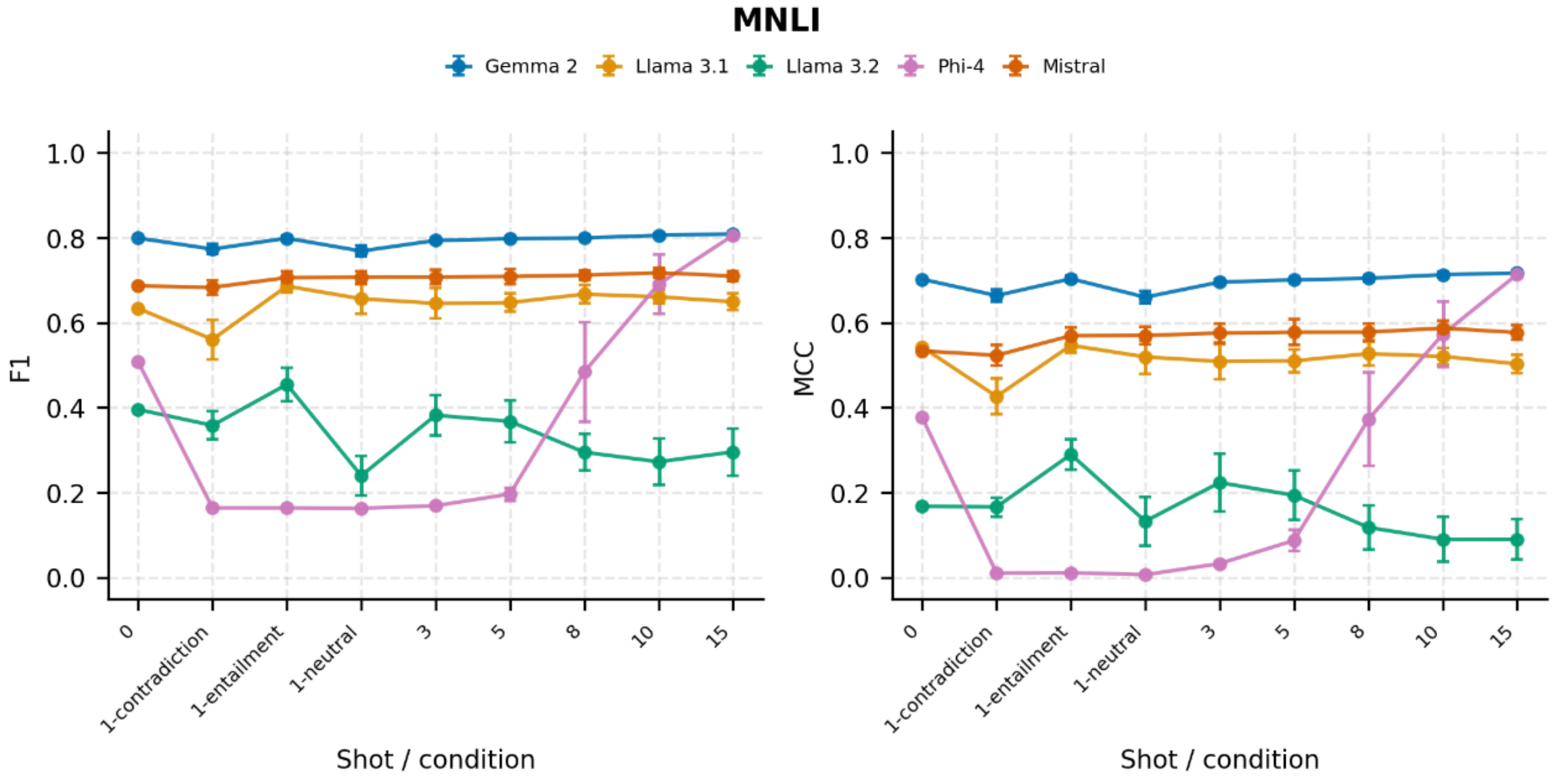

(a) F1 (b) MCC

**Fig. 2** Performance of each model on MNLI as a function of the number of demonstration examples in the prompt. (a) F1-score and (b) MCC

Figure 2 shows the results on MNLI, with F1-score in panel (a) and MCC in panel (b). Gemma 2, Llama 3.1 and Mistral performed very consistently across shot counts. Phi-4, as discussed in section 2, collapsed for up to five demonstration examples, and at eight its confidence interval was quite wide. With more examples, however, it reached the performance of Gemma 2. Lastly, Llama 3.2 was very inconsistent with wide confidence intervals, and underperformed at all shot counts. Since MNLI is balanced across classes, MCC follows the same trends as F1-score, and both panels show the same patterns.

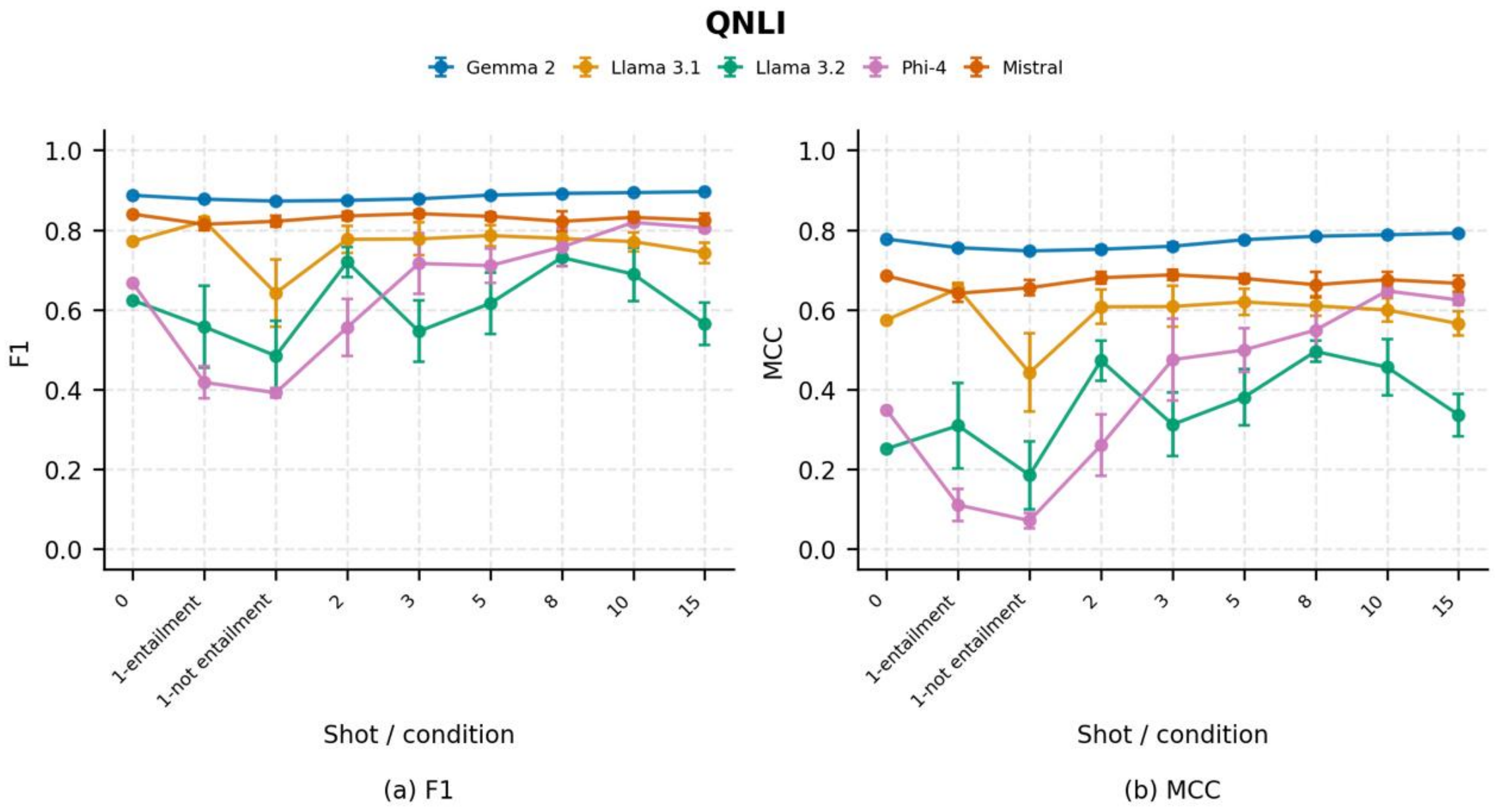

**Fig. 3** Performance of each model on QNLI as a function of the number of demonstration examples in the prompt. (a) F1-score and (b) MCC

Figure 3 shows the results on QNLI, with F1-score in panel (a) and MCC in panel (b). Here the picture is slightly different: most models performed very well, and Phi-4 did not collapse with few examples as it did on MNLI. Llama 3.2, however, remained inconsistent and unpredictable, with wide confidence intervals. It is also worth noting that Llama 3.1's performance only declined as examples were added, apart from the single entailment example. Since QNLI is also balanced, the trends between MCC and F1-score are very similar.

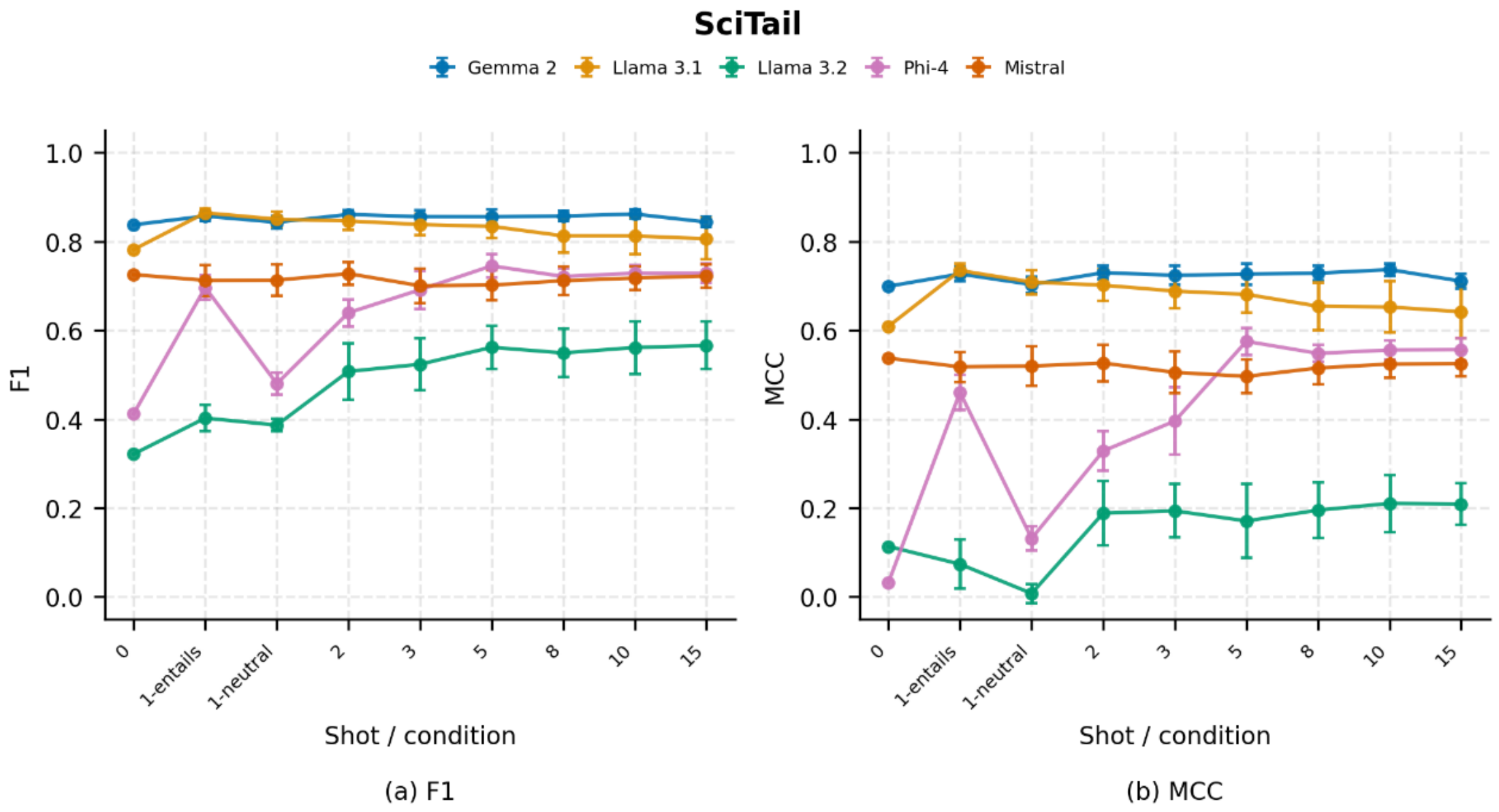

**Fig. 4** Performance of each model on SciTail as a function of the number of demonstration examples in the prompt. (a) F1-score and (b) MCC

Figure 4 shows the results on SciTail, with F1-score in panel (a) and MCC in panel (b). SciTail is an interesting case, as all models show clear patterns: Gemma 2, Llama 3.1 and Mistral are practically flat, while Phi-4 and Llama 3.2 improve as the number of examples increases. All models also show wider confidence intervals than on the other datasets, suggesting that performance here is more sensitive to which demonstration examples are selected. Unlike MNLI and QNLI, SciTail is not balanced, so MCC shows some noticeable shifts compared to F1-score, although the general trend remains the same. One exception is Llama 3.2, where F1-score gradually increases as the number of examples increases but MCC drops for both one-shot conditions.

## *5.4 Discussion*

In Table 7 we present a side-by-side performance comparison between our results, different model architectures, and similar studies based on the MNLI dataset. Alongside the results reported in Section V there is a very strong indication that small models are still not able to compete with encoder-only models or encoder-decoder. We can see that the T5 model is outperforming the rest of the field (+11.36% accuracy from Gemma 2 9B), with DeBERTa being a close second (+10.24% accuracy from Gemma 2 9B). However, in our efforts to create a baseline for decoder-only models on well-established datasets we found that our approach shows very promising results when compared to similar evaluations. For Mistral on the MNLI dataset our setup scored a higher accuracy of 75.01% when compared to ICS [12] and MCQs prompting [30]. Similarly for Llama 3.1 8B, our approach had a better performance (68.68% accuracy) compared to the MCQs prompts from [30]. This finding shows that there is still more research needed in the field of prompt engineering and that there is still room for improvement in order to reach the performance of state-of-the-art encoder-decoder models.

**Table 7** Comparison of results between model architectures and techniques on MNLI Dataset. Bold denotes our results

| Model | Model Type | Accuracy |
|---|---|---|
| **Llama 3.2 3B (Instruct)** | **Decoder-only** | **0.504** |
| ***Llama 3.1 8B (Instruct)*** | **Decoder-only** | **0.689** |
| *Llama 3.1 8B (Base-MCQ) [30]* | Decoder-only | *~0.6700* |
| ***Gemma 2 9B (Instruct)*** | **Decoder-only** | **0.8086** |
| ***Mistral 7B (Instruct)*** | **Decoder-only** | **0.722** |
| ***Phi 4 Mini 3.8B (Instruct)*** | **Decoder-only** | **0.8074** |
| *Mistral 7B (Base-MCQ) [30]* | Decoder-only | *~0.6700* |
| *Mistral 7B (Instruct-ICS) [12]* | Decoder-only | 0.6340 |
| *Flan - T5 XL (ICS) [12])* | Encoder-decoder | 0.7010 |
| *T5 [28]* | Encoder-decoder | 0.9220 |
| *DeBERTa [7]* | Encoder-only | 0.9110 |

There is also a clear correlation between the size of the model and its performance, which is in accordance with recent studies [30, 36]. This is directly visible from our results in Table 7 where Gemma 2 9B reached an accuracy of 80.86% whereas Llama 3.2 3B achieved only 50.4%. Gemma 2 clearly outperformed the rest of the field apart from the SciTail dataset as shown in Table 2 and Figure 4, where it came second only by a small margin. This is likely attributed to differences in the knowledge domains when the models were trained. A strong exception is Phi-4, which shows very strong predictive capabilities, however it seems that it needs many examples in order to perform as well as other models. Finally, our findings strongly support our claim that a few demonstration examples and efficient prompting strategies are able to help the models reason better than many examples or highly complex prompting techniques, while maintaining low computational cost.

One possible explanation for the effectiveness of simple prompts is that they reduce the cognitive load imposed on the model during in-context processing. As the number of demonstration examples increases, the model must allocate attention across a larger prompt context while simultaneously identifying the task-relevant semantic relationship between premise and hypothesis. This additional contextual complexity may introduce noise and interfere with the model's ability to maintain stable decision boundaries. This stands particularly for smaller models with limited representational capacity.

**Table 8** Comparison of F1-Score between best result and zero-shot setting. Effect size is the rank-biserial

| Model | Dataset | F1-Score Zero-Shot | F1-Score best | p-value | Effect Size r |
|---|---|---|---|---|---|
| Llama 3.2 | QNLI | 0.6240 | 0.731 ± 0.022 | .005 | 0.89 |
| | MNLI | 0.3950 | 0.454 ± 0.040 | .010 | 0.79 |
| | SciTail | 0.3220 | 0.566 ± 0.054 | .005 | 0.89 |
| *Llama 3.1* | QNLI | 0.7710 | 0.822 ± 0.010 | .005 | 0.89 |
| | MNLI | 0.6340 | 0.686 ± 0.014 | .005 | 0.89 |
| | SciTail | 0.7820 | 0.865 ± 0.010 | .005 | 0.89 |
| *Mistral* | QNLI | 0.8397 | 0.8406 ± 0.0083 | **.300** | **0.31** |
| | MNLI | 0.6870 | 0.718 ± 0.011 | .005 | 0.89 |
| | SciTail | 0.7250 | 0.728 ± 0.025 | **.900** | **0.05** |
| *Gemma 2* | QNLI | 0.8868 | 0.8959 ± 0.0017 | .005 | 0.89 |
| | MNLI | 0.7993 | 0.8089 ± 0.0046 | .007 | 0.85 |
| | SciTail | 0.8377 | 0.8618 ± 0.0097 | .005 | 0.89 |
| *Phi 4* | QNLI | 0.6664 | 0.8189 ± 0.0077 | .005 | 0.89 |
| | MNLI | 0.5082 | 0.8059 ± 0.0047 | .005 | 0.89 |
| | SciTail | 0.4120 | 0.745 ± 0.026 | .005 | 0.89 |

In Table 8 we report the results of zero-shot as a baseline and the best for each model-dataset combination and compare them to see if the differences found are statistically significant. The results indicate a clear trend; few-shot is helping with the performance of the models. One exception however is the Mistral model, which shows very strong capabilities in zero-shot settings.

## 6. Limitations

In our study we have not examined the effect of demonstration examples permutations. A recent study shows that the order with which the examples are shown in few-shot prompting has a real effect on the performance of the model [37]. While future work could investigate this finding, we believe that it is out of scope for our current work. Moreover, we strongly believe that certain sampling or prompting techniques, such as Chain-of-Thought [14] or Metacognitive Prompting [13], can potentially yield better results but the main focus of this study was to benchmark popular LLMs and present a baseline for future studies.

During the experimentation we noted that minor variations in the phrasing of the prompt had a small but measurable effect on evaluation results. This observation highlights the sensitivity of modern LLMs to prompt wording and indicates that prompt standardization remains important for evaluation.

Moreover, while the datasets chosen are representative of different knowledge domains and have various difficulty, they do not account for other specialized domains such as medical NLI, where the models could be challenged differently.

Lastly, all decoder-only models were evaluated using 8-bit quantization. This choice reduces memory requirements. However, it may introduce small differences compared with full-precision inference.

## 7. Conclusions

This study presented a comprehensive evaluation of decoder-only large language models on traditional Natural Language Inference benchmarks. The results demonstrate that models such as Gemma can achieve performance comparable to encoder-based architectures when guided by simple and direct prompting strategies, although there is a real gap between decoder-only models and encoder-only or encoder-decoder. In contrast, smaller models, such as Llama, exhibit significantly lower performance, highlighting the impact of model scale on inference capabilities. At the same time, the results indicate that carefully designed simple prompting strategies can substantially improve performance without requiring complex and reasoning-oriented prompt engineering methods.

A key finding of this study is the effect of the demonstration example conditions in one-shot settings, suggesting that label selection, rather than example selection strategies is an important factor in single-shot NLI performance. Although the reason behind is not clear, we hypothesize that entailment examples provide more useful information that then help the models differentiate between the labels.

Additionally, this work shows that prompt simplicity plays a crucial role in model performance. In both zero-shot and few-shot settings, simple prompts allow models to focus more effectively on the task, while the use of a small number of representative examples often yields better results than larger, more complex prompt configurations. Moreover, performance was found to depend on the interaction between model architecture, dataset characteristics, and prompt design. These findings emphasize the importance of benchmarking LLMs on traditional NLI datasets, particularly for smaller and more efficient models. Such evaluations provide valuable insights into their reasoning capabilities and practical applicability.

Regarding our exact research questions, the findings show that modern instruction-tuned decoder-only LLMs, particularly Gemma 2 9B, can approach encoder-based performance on NLI tasks, but there is a clear gap, especially for small models (RQ1). In addition, the label of the single demonstration example significantly affected model predictions, and reveals that the choice of label class is a key design variable in single-shot Natural Language Inference (RQ2). Furthermore, increasing the number of in-context demonstration examples beyond a small set, generally did not provide further gains and in several cases degraded performance, something that indicates that prompt efficiency is more beneficial than prompt volume (RQ3). Moreover, it is important to note that our simple, direct prompting strategy consistently matched or outperformed more elaborate techniques reported in the literature, such as ICS and MCQ-based prompting (RQ4).

Building on the present findings, future work will focus on some main directions. First, we want to investigate the reasons why entailment-primed examples have this effect, and whether it can be found in other NLP tasks. Secondly, future work will explore and perform deeper comparisons between simple prompting and structured reasoning strategies — such as Chain-of-Thought, self-consistency, and metacognitive prompting — to further examine their impact on NLI tasks. Future work will also examine different prompt formulations demonstration orders and label verbalisers. Another direction is to draw demonstrations exclusively from training partitions, so that selection is fully separated from evaluation data. Finally, evaluating on additional domain-specific and multilingual NLI datasets could help assess the generality of the findings.

## **APPENDIX A -** Confusion Matrices

In this section we present the confusion matrices for zero-shot and for the entailment-primed one-shot cases that were found statistically significant.

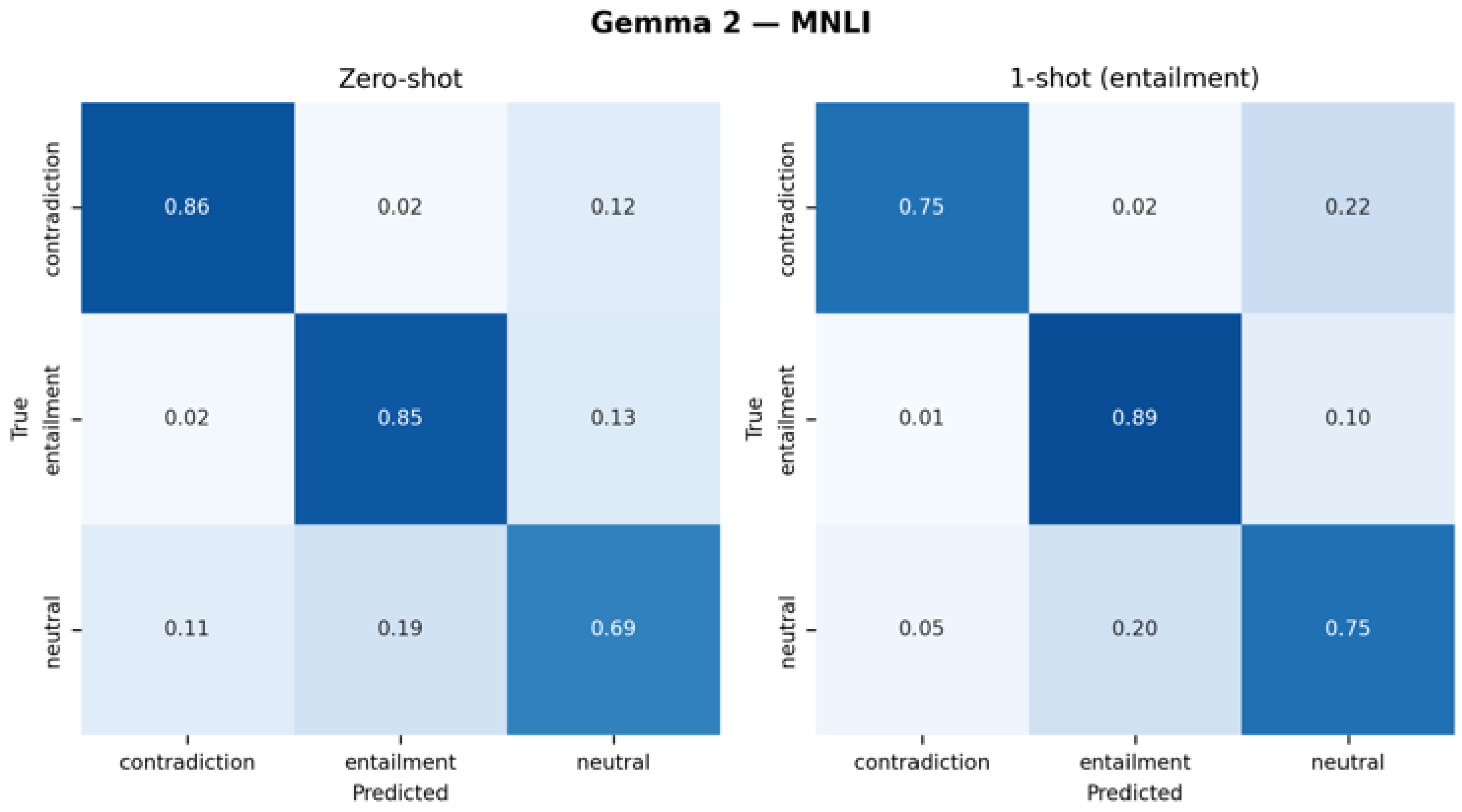

**Fig. 5** Confusion matrices (row-normalized) for Gemma 2 on MNLI, comparing zero-shot and one-shot predictions when primed with an entailment demonstration example.

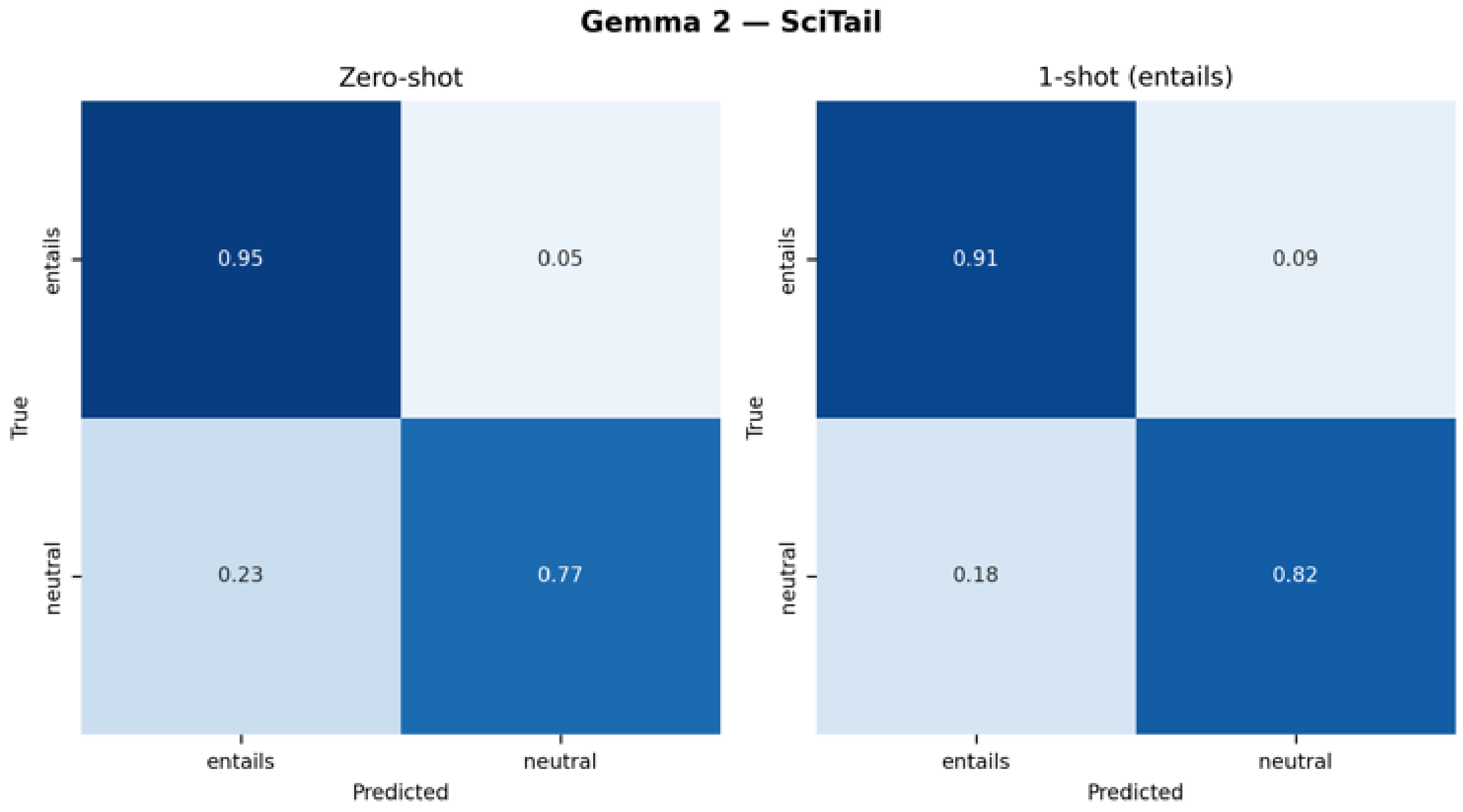

**Fig. 6** Confusion matrices (row-normalized) for Gemma 2 on SciTail, comparing zero-shot and one-shot predictions when primed with an entailment demonstration example.

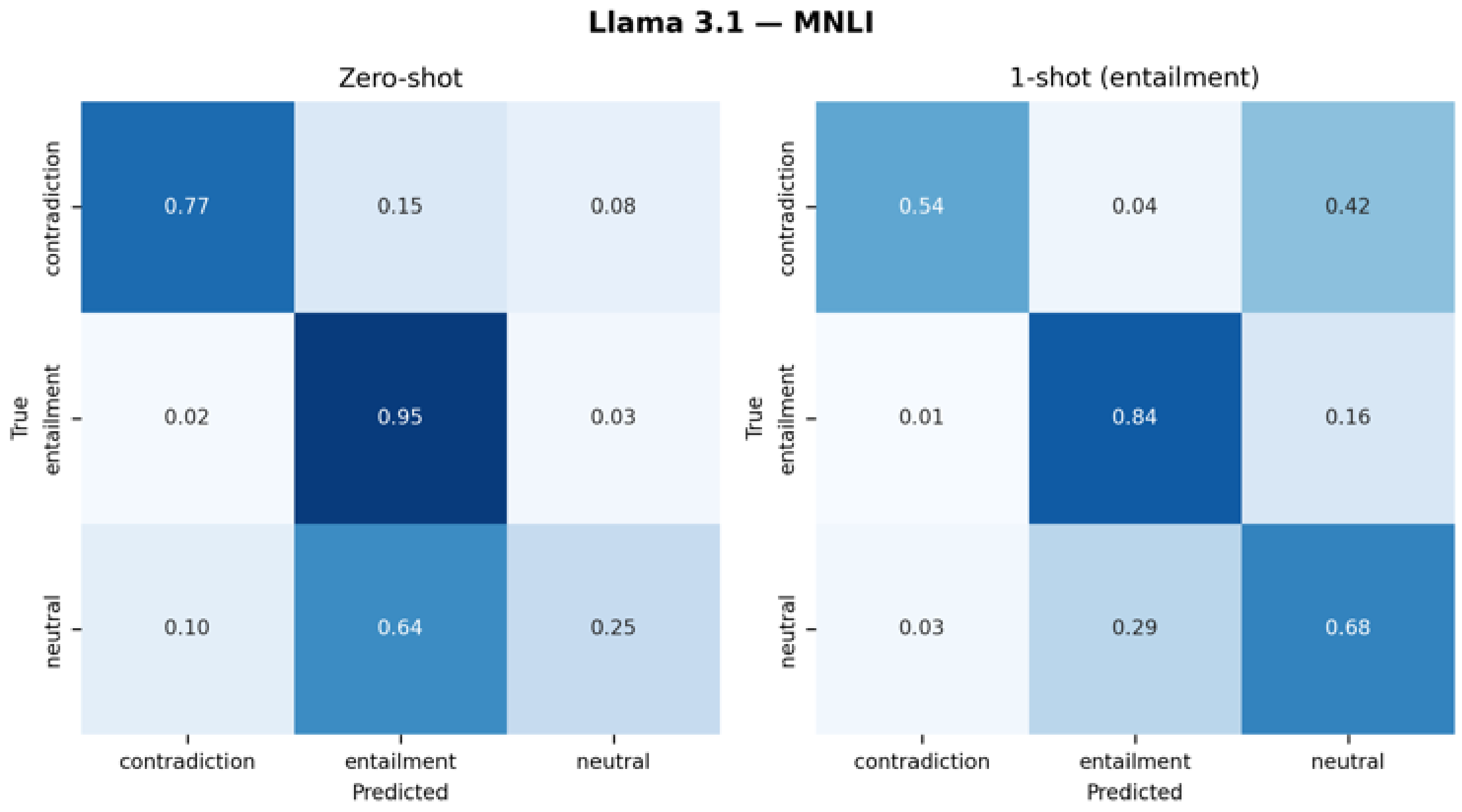

**Fig. 7** Confusion matrices (row-normalized) for Llama 3.1 on MNLI, comparing zero-shot and one-shot predictions when primed with an entailment demonstration example.

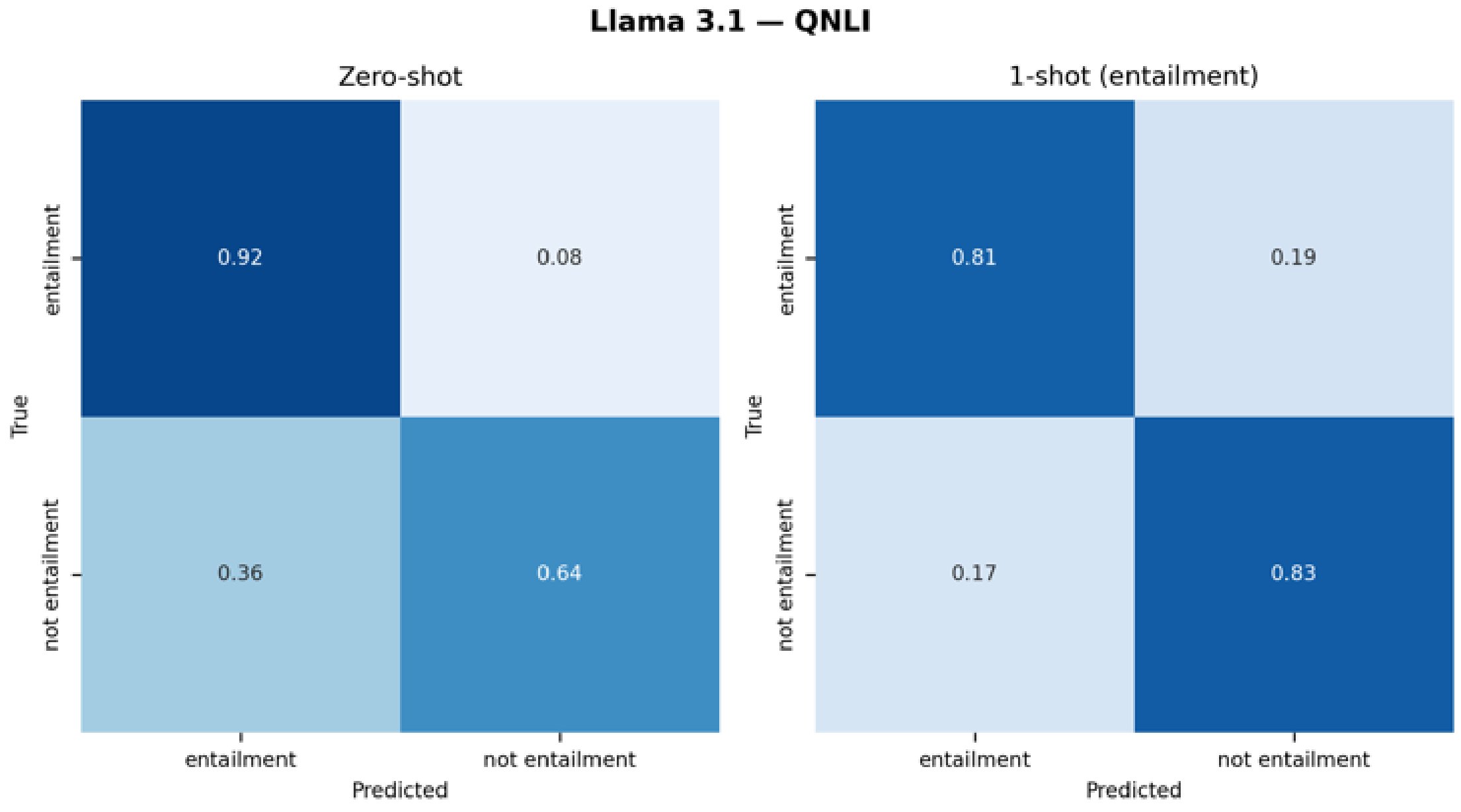

**Fig. 8** Confusion matrices (row-normalized) for Llama 3.1 on QNLI, comparing zero-shot and one-shot predictions when primed with an entailment demonstration example.

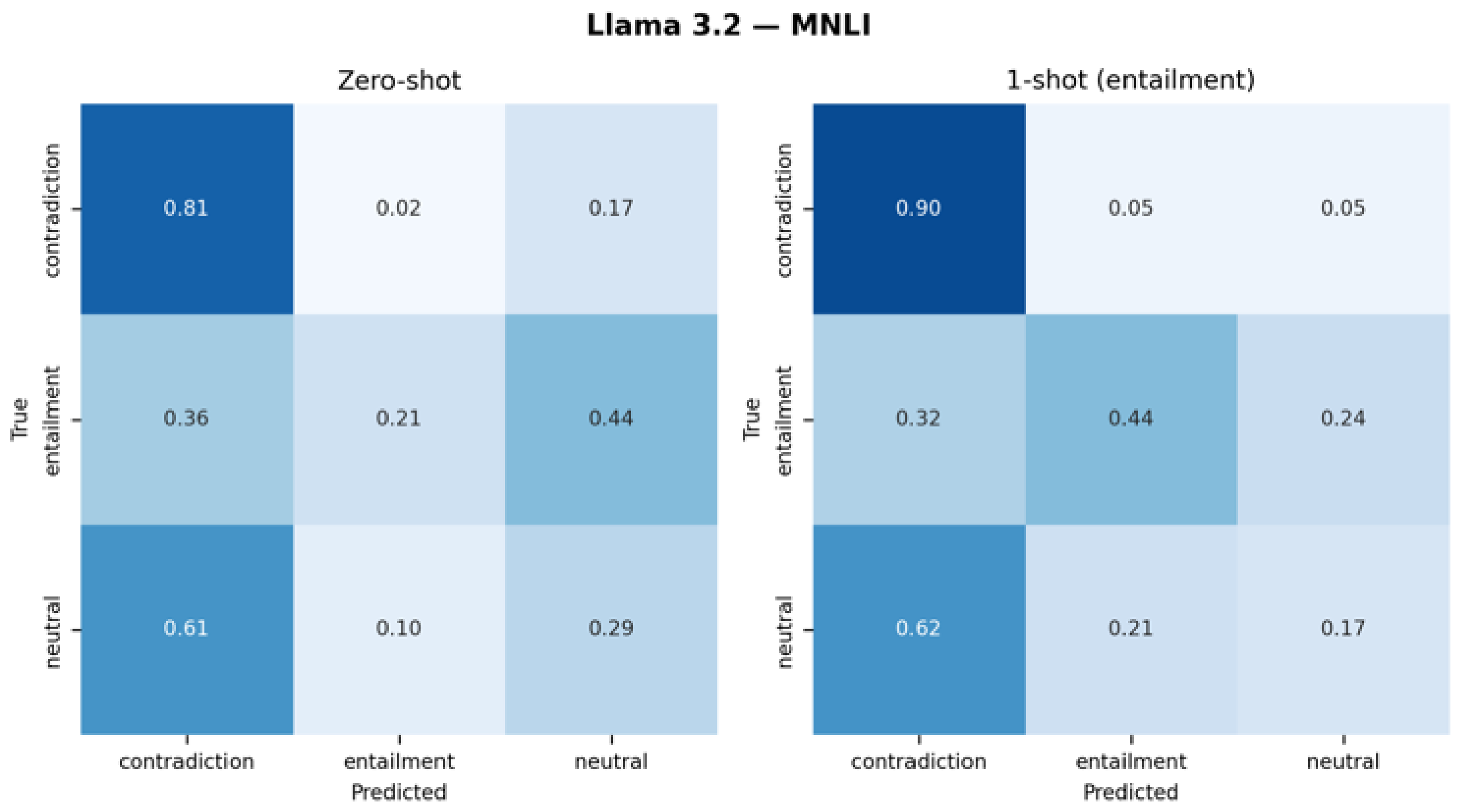

**Fig. 9** Confusion matrices (row-normalized) for Llama 3.2 on MNLI, comparing zero-shot and one-shot predictions when primed with an entailment demonstration example.

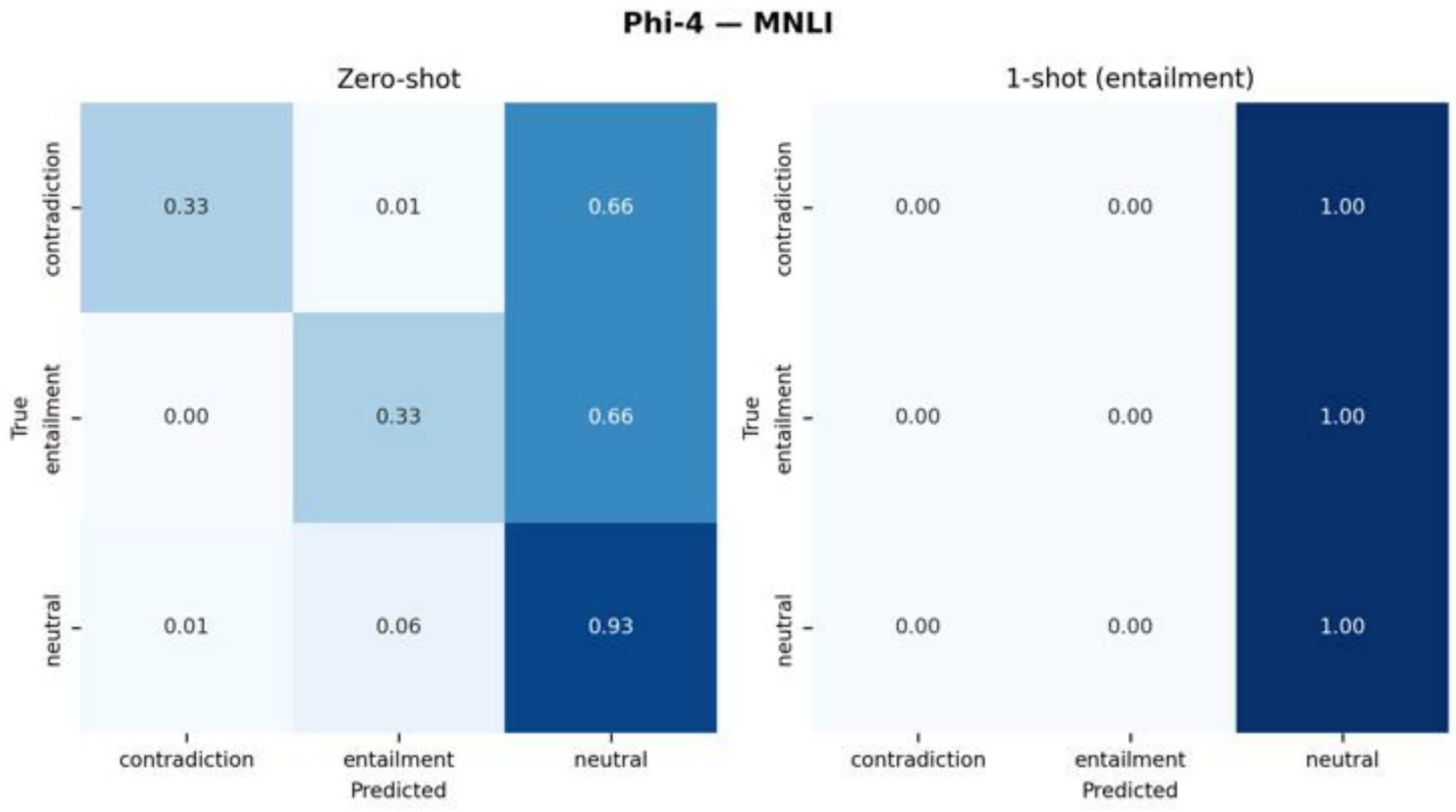

**Fig. 10** Confusion matrices (row-normalized) for Phi-4 on MNLI, comparing zero-shot and one-shot predictions when primed with an entailment demonstration example.